\documentclass[twoside,11pt]{article}
\usepackage{jmlr2e}
\usepackage{amsmath,amssymb}
\ShortHeadings{Locally Coupled Gaussian Process Analysis}{Ambrogioni et al}

\begin{document}
\vspace*{0.35in}

\begin{flushleft}
{\Large
\textbf\newline{Analysis of Nonstationary Time Series Using Locally Coupled Gaussian Processes}
%\textbf\newline{Isolating Neural Processes via Spatiotemporal GP Decomposition}
}
\newline
\\
Luca Ambrogioni\textsuperscript{1} and
Eric Maris\textsuperscript{1}
\\
\bigskip
\bf{1} Radboud University, Donders Institute for Brain, Cognition and Behaviour, Nijmegen, The Netherlands 
\\
\bigskip
* l.ambrogioni@donders.ru.nl

\end{flushleft}

\begin{abstract}
The analysis of nonstationary time series is of great importance in many scientific fields such as physics and neuroscience. In recent years, Gaussian process regression has attracted substantial attention as a robust and powerful method for analyzing time series. In this paper, we introduce a new framework for analyzing nonstationary time series using locally stationary Gaussian process analysis with parameters that are coupled through a hidden Markov model. The main advantage of this framework is that arbitrary complex nonstationary covariance functions can be obtained by combining simpler stationary building blocks whose hidden parameters can be estimated in closed-form. We demonstrate the flexibility of the method by analyzing two examples of synthetic nonstationary signals: oscillations with time varying frequency and time series with two dynamical states. Finally, we report an example application on real magnetoencephalographic measurements of brain activity.
\end{abstract}


\section{Introduction}

Time series are often nonstationary. Nevertheless, many popular methods for time series analysis assume stationarity of the underlying process. For example, this holds for many applications of Gaussian process (GP) regression. GP regression provides an intuitive and flexible way of analyzing time series using a non-parametric prior model of their dynamics, as expressed by the covariance function \citep{rasmussen2006gaussian, paciorek2004nonstationary, hartikainen2010kalman}. Typically this prior covariance function is assumed to be stationary, i.e. solely dependent on the difference between time points. Using nonstationary covariance functions can be challenging since the nature of the nonstationarity is often not known a priori and can be difficult to estimate from the data. 

A possible approach is to define a hierarchical model where the covariance function depends on time dependent parameters. The posterior distribution of these parameters can be estimated by MCMC sampling \citep{paciorek2004nonstationary} or locally optimized using a gradient based search \citep{plagemann2008nonstationary, kersting2007most}. Another possibility is to extend the input space with unobservable variables and use a covariance function that is stationary in this expanded space \citep{pfingsten2006nonstationary}. Recently, some authors introduced nonstationary and heteroscedastic models by using GP hyper priors for the noise and signal variances \citep{munoz2011heteroscedastic, tolvanen2014expectation}. 

All these approaches are not very flexible as they do not allow full freedom in specifying the nature of the non-stationarity. In general, it is difficult to define a flexible family of parametrized covariance functions that can model different kinds of non-stationarities due to the requirement of positive definiteness across the parameter space. A more flexible path is to partition the time series into small segments, each modeled by a local stationary GP \citep{kim2005analyzing, gramacy2008bayesian}. In this paper, we follow a somewhat similar strategy by modeling nonstationary time series using a positive linear mixture of local GPs whose parameters are coupled through a hidden Markov model \citep{rabiner1986introduction}. We called this new method \emph{locally coupled GP regression}. Since a positive linear mixture of GPs is always a valid GP, we are free to use any combination of stationary covariance functions with arbitrary local coupling of their parameters. When these parameters have a finite range, the inference can be performed exactly using the forward-backward algorithm \citep{forney1996forward}. This approach allows to combine the flexibility of partition based methods with the regularization properties of hierarchical models (that usually assume smoothness in the hyper-parameters) in a common, analytically solvable, framework.

In this paper we do not aim to systematically compare the performance of the existing specialized methods for nonstationary GP regression. Rather, we demonstrate the flexibility and robustness of our new method and give some examples of its range of applications.

\subsection{Gaussian Process Regression}
In this subsection, we review the concept of GP and its relevance for machine learning and time series analysis. GPs are the infinite dimensional generalization of multivariate Gaussian distributions. They are fully specified by their mean and covariance functions, the analogues of respectively the mean vector and the covariance matrix in the multivariate Gaussian. In the machine learning literature, the mean function is often assumed to be identically equal to zero \citep{rasmussen2006gaussian}. The covariance function $k(t,t';\boldsymbol{\vartheta})$ defines the prior covariance between the values of the process at different time points. It has to be symmetric and non-negative definite, meaning that all its eigenvalues are non-negative. In the rest of the paper, we will parameterize the covariance function with a vector of hyper-parameters $\boldsymbol{\vartheta}$.

GPs can be used as non-parametric prior distributions over the space of continuous-time signals. Given a set of $n$ sample points $T = \{t_1,....t_n\}$, we can model a discretely measured time series $\boldsymbol{s}$ as follows:
\begin{equation}
s_j = \psi(t_j) + \epsilon_j~,
\label{observation model GP, introduction}
\end{equation}
where $\epsilon_j$ is white noise process with variance $\lambda$ and the continuous time latent process $\psi(t)$ follows a GP distribution with mean $0$ and covariance function $k_\psi(t, t'; \boldsymbol{\vartheta})$
\begin{equation}
\psi(t_j) \sim GP\big(0, k_\psi(t, t'; \boldsymbol{\vartheta})\big)~.
\label{GP distribution, introduction}
\end{equation}
While we can obtain the posterior distribution of $\psi(t)$ at any arbitrary set of target points, in this paper we will restrict our attention to the sample point themselves. To this end, we introduce the vector $\boldsymbol{\psi}$ with components $\psi_j = \psi(t_j)$. As both the prior and the likelihood are multivariate Gaussian distributions, the posterior distribution $p(\boldsymbol{\psi}|s, \boldsymbol{\vartheta})$ is also a multivariate Gaussian. In particular, its expected value $\boldsymbol{m}_{\psi|s}$ is given by \citep{rasmussen2006gaussian}
\begin{equation}
\boldsymbol{m}_{\psi|s, \boldsymbol{\vartheta}} = K_\psi(\boldsymbol{\vartheta}) (K_\psi(\boldsymbol{\vartheta}) + \lambda I)^{-1} \boldsymbol{s},
\label{posterior expectation GP, introduction}
\end{equation}
where $\lambda I$ is the covariance matrix of the white noise and the the $i,j$-th entry of the sample points covariance matrix $K_\psi(\boldsymbol{\vartheta})$ is $k_\psi(t_j,t_k; \boldsymbol{\vartheta})$. 
The hyper-parameters $\boldsymbol{\vartheta}$ can be estimated hierarchically by introducing a hyper-prior $p(\boldsymbol{\vartheta})$ and marginalizing the latent process 
\begin{equation}
p(\boldsymbol{\vartheta}|\boldsymbol{s}) \propto p(\boldsymbol{\vartheta}) \int p(\boldsymbol{s}|\boldsymbol{\psi}) p(\boldsymbol{\psi}|s, \boldsymbol{\vartheta}) d\boldsymbol{\psi} = p(\boldsymbol{\vartheta}) p(\boldsymbol{s}|\boldsymbol{\vartheta})~.
\label{posterior hyper-parameters, introduction}
\end{equation}
The marginal likelihood $p(\boldsymbol{s}|\boldsymbol{\vartheta})$ can be obtained analytically, its logarithm is given by
\begin{equation}
\log p(\boldsymbol{s}|\boldsymbol{\vartheta}) = -\frac{n}{2}\log 2 \pi -\frac{1}{2} \log |K_\psi(\boldsymbol{\vartheta})  + \lambda I|  - \frac{1}{2} \boldsymbol{s}^T \big(K_\psi(\boldsymbol{\vartheta})  + \lambda I \big)^{-1} \boldsymbol{s}~.
\label{marginal likelihood, introduction}
\end{equation}
In most interesting situations the prior $p(\boldsymbol{\vartheta})$ is not conjugate to the marginal likelihood. Therefore, the normalization integral is often approximated using methods such as grid integration or MCMC \citep{rasmussen2006gaussian}.

\subsection{Stationary and Nonstationary Covariance Functions}
The smoothing and regularization properties of GP regression depend on the choice of the covariance function. A covariance function $k(t,t')$ is said to be stationary when it solely depends on the difference between the time points $t' - t$. A commonly used example of a stationary covariance function is the squared exponential
\begin{equation}
k_{SE}(t, t'; l) = e^{-\frac{(t - t')^2}{2 \delta^2}}~,
\label{squared exponential, methods}
\end{equation}
where the parameter $\delta$ regulates the time scale of the process. Intuitively, this is the time you need to wait before the process value will significantly change. The squared exponential covariance is simple and intuitive but it assumes that the length scale of the process does not change as a function of time. Another example of stationary covariance is given by the oscillatory covariance function:
\begin{equation}
k_{O}(t, t'; l) =e^{-\frac{(t - t')^2}{2 d^2}} \cos{\omega (t - t')}~.
\label{oscillatory, methods}
\end{equation}
This covariance function generates stochastic oscillations whose angular frequency peaks at $\omega$. The parameter $d$ specifies the spectral concentration of the oscillations around $\omega$.

Several families of nonstationary covariance functions have been proposed (for example see \citep{williams1998computation, gibbs1998bayesian, paciorek2004nonstationary}). In this paper, we define a locally stationary GP as a weighted sum of independent stationary GPs. Each stationary GP, here denoted as $\phi_i(t)$, models a short segment of the time series around the support point $t_i$. The global nonstationary process $\zeta(t)$ is a linear mixture of these local processes: 
\begin{equation}
\zeta(t) = \sum_{i} w(t;t_i) \phi_i(t)~,
\label{locally stationary GP, methods}
\end{equation}
where $w(t;t_i)$ is a basis function localized around the support point $t_i$ and $\phi_j(t)$ is a GP. A linear combination of GPs is always a GP itself, its covariance function is given by
\begin{equation}
k_\zeta(t,t';\{\boldsymbol{\vartheta}_1,...\}) = \sum_{i,j}  w(t;t_i) w(t';t_j) \langle \phi_i(t) \phi_j(t') \rangle =  \sum_j  w(t;t_i) w(t';t_i) k_{i}(t,t';\boldsymbol{\vartheta}_i)~,
\label{locally stationary GP covariance, methods}
\end{equation}
where $\{\boldsymbol{\vartheta}_1,...\}$ denotes a set of vector parameters, one vector for each local process. In this expression we denoted the covariance function of $\phi_i(t)$ as $k_{i}(t,t';\boldsymbol{\vartheta}_i)$. In the rest of the paper we will work with a regular grid of support points. In addition, we will normalize the basis functions in order to not induce prior non-stationarity in the variance
\begin{equation}
\sum_i w(t;t_i)^2 = 1~.
\label{basis function normalization, methods}
\end{equation}
The family of nonstationary covariance functions given by Eq.~\ref{locally stationary GP covariance, methods} is very flexible, as we are free to specify any stationary or nonstationary covariance functions $k_{i}(t,t';\boldsymbol{\vartheta}_i)$ without breaking the positive definiteness of $k_\zeta(t,t';\{\boldsymbol{\vartheta}_1,...\})$.

\subsection{Locally Coupled Gaussian Process Regression}
We will now explain how to estimate the local hyper-parameters of the model. Estimating these hyper-parameters allows to infer the structure of the nonstationary directly from the data. We can then use these estimates for constructing a global nonstationary covariance function from Eq.~\ref{locally stationary GP covariance, methods} that can in turn be used for analyzing the whole time series.

The covariance function given by Eq.~\ref{locally stationary GP covariance, methods} depends on a family of vector hyper-parameters $\{\boldsymbol{\vartheta}_1,...\}$. The main idea of this paper is to learn these parameters by considering the vector variables $\{\boldsymbol{\vartheta}_1,...\}$ as an hidden Markov model \citep{rabiner1986introduction}. This induces a local coupling between the parameters of neighboring local GPs and henceforth smooths the estimation of the parameters. For example, we can use an autoregressive process
\begin{equation}
\boldsymbol{\vartheta_{k+1}} = A \boldsymbol{\vartheta_k} + \boldsymbol{\xi}~,
\label{hidden autoregressive model, methods}
\end{equation}
where $A$ is a matrix of autoregressive coefficients and $\boldsymbol{\xi}$ is a centered normal random variable. Once we specify an initial probability distribution for $\boldsymbol{\vartheta}_0$, this model defines a prior distribution over $\{\boldsymbol{\vartheta}_1,...\}$. The likelihood of the model is obtained by marginalizing the $\zeta(t)$ using Eq.~\ref{marginal likelihood, introduction} (where $\psi(t)$ should be replaced by $\zeta(t)$). In particular, if the basis functions $w_i(t)$ have disjoint supports, each $\boldsymbol{\vartheta}_i$ solely affects a segment of time series $\boldsymbol{s}_i$. In this case, the vector valued observations $\boldsymbol{s}_i$ are conditionally independent given the hidden variables $\{\boldsymbol{\vartheta}_1,...\}$ and, consequently, the marginal likelihood $p(\boldsymbol{s}|\{\boldsymbol{\vartheta}_1,...\})$ factorizes as follows
\begin{equation}
p(\boldsymbol{s}|\{\boldsymbol{\vartheta}_1,...\}) = \prod_i p(\boldsymbol{s}_i|\boldsymbol{\vartheta}_i)
\label{dynamic marginal likelihood, methods}
\end{equation}
in which the segment specific marginal likelihood $p(\boldsymbol{s}_i|\boldsymbol{\vartheta}_i)$ is obtained from Eq.~\ref{marginal likelihood, introduction} using the local covariance matrix
\begin{equation}
K_i(\boldsymbol{\vartheta}_i) = (\boldsymbol{w}^{(i)}{\boldsymbol{w}^{(i)}}^T)\odot K_i(\boldsymbol{\vartheta}_i)
\label{segment covariance matrix, methods}
\end{equation}
where the basis vectors $\boldsymbol{w}^{(i)}$ are defined by the entries $w^{(i)}_j = w(t_j;t_i)$ and $\odot$ denotes the entriwise product between matrices. It is easy to check that the covariance matrix given by Eq.~\ref{segment covariance matrix, methods} is induced by the covariance function $w(t;t_i) w(t';t_i) k_{i}(t,t';\boldsymbol{\vartheta}_i)$.

Summarizing, the vector valued observations $\boldsymbol{s}_i$ are conditionally independent given the hidden variables $\{\boldsymbol{\vartheta}_1,...\}$, which in turn form a Markov chain. The posterior of this kind of Bayesian networks can be obtained by recursive Bayesian estimation \citep{sarkka2013bayesian}. This estimation can be performed exactly when the system is linear and Gaussian using the Kalman filter \citep{kalman1960new} or when the variables $\{\boldsymbol{\vartheta}_1,...\}$ have a finite range using the Forward-Backward algorithm \citep{forney1996forward}. Otherwise, the estimation can be approximated using methods such as the extended Kalman filter or a particle filter \citep{sarkka2013bayesian}. 

Using one of these methods, we can obtain the posterior distributions $p(\boldsymbol{\vartheta}_i |\boldsymbol{s}_i)$ for each local GP. From the posterior, we can extract point estimates $\hat{\boldsymbol{\vartheta}}_i$ and use them as parameters of the GP regression determined by Eq.~\ref{posterior expectation GP, introduction} and \ref{locally stationary GP covariance, methods}. For example, we can use the posterior mean as point estimate:
\begin{equation}
\hat{\boldsymbol{\vartheta}}_i = \int \boldsymbol{\vartheta}_i p(\boldsymbol{\vartheta}_i |\boldsymbol{s}_i) d\boldsymbol{\vartheta}_i~.
\end{equation}
Using the point estimates of the hyper-parameters, we can construct the global nonstationary covariance function from Eq.~\ref{locally stationary GP covariance, methods}. This covariance function can subsequently be used for analyzing the time series.

The requirement that the basis functions $w(t;t_i)$ have disjoint supports is quite restrictive since it completely disconnects the GP regression between the segments of data. Therefore, in the following we will remove this constraint but we will still approximate the marginal likelihood using Eq.~\ref{dynamic marginal likelihood, methods}. This approximation entails a loss of temporal resolution since each local process will attempt to explain some features of the time series that are already explained by either the previous or the next local process. Note that, when we analyze the data using the resulting global covariance function, we do not need to make this assumption. Therefore, since each local GP is itself a flexible non-parametric model, the second step is likely to correct for most of the distortions introduced by approximating the marginal likelihood of the hyper-parameters.

\section{Experiments}
In the following we show some applications of the locally coupled GP analyses. We start with two analysis on simulated data, one with an oscillatory signal with variable frequency and the other with a signal that switches between two dynamical states. Finally, we show how the two-state model can be used for identifying bursts of alpha oscillations in the magnetic brain signal, as measured by MEG.

\subsection{Analysis of Nonstationary Oscillations}
The instantaneous frequency of many real world oscillatory signals of great scientific interest changes as a function of time \citep{atallah2009instantaneous, roberts2013robust, baker2006gravitational}. Stationary filters, including stationary GP regressions, cannot adjust their gain to these frequency shifts and are therefore sub-optimal for this kind of signals. 

Here, we use the locally coupled GP regression in order to construct a filter that can track the frequency shifts and change its gain consequently. Each local stationary GP is equipped with the oscillatory covariance function in Eq.~\ref{oscillatory, methods}. In this case, the hidden states are scalar variables that determine the local angular frequencies $\omega_i$, which we discretized in a range from $0.1$ Hz to $12$ Hz in steps of $0.4$ Hz. The transition probabilities between frequencies were obtained from the continuous autoregressive model given by Eq.~\ref{hidden autoregressive model, methods} (with $A$ equal to $1$ and $\langle \xi^2 \rangle$ equal to $0.2^2$) and subsequently  renormalized in order to account for the discretization and the finiteness of the range. We used Gaussian window functions with width (standard deviation) equal to $0.25$ s and spaced by $0.1$ s.

\begin{figure}
	\centering
    	\includegraphics[width=0.9\textwidth] {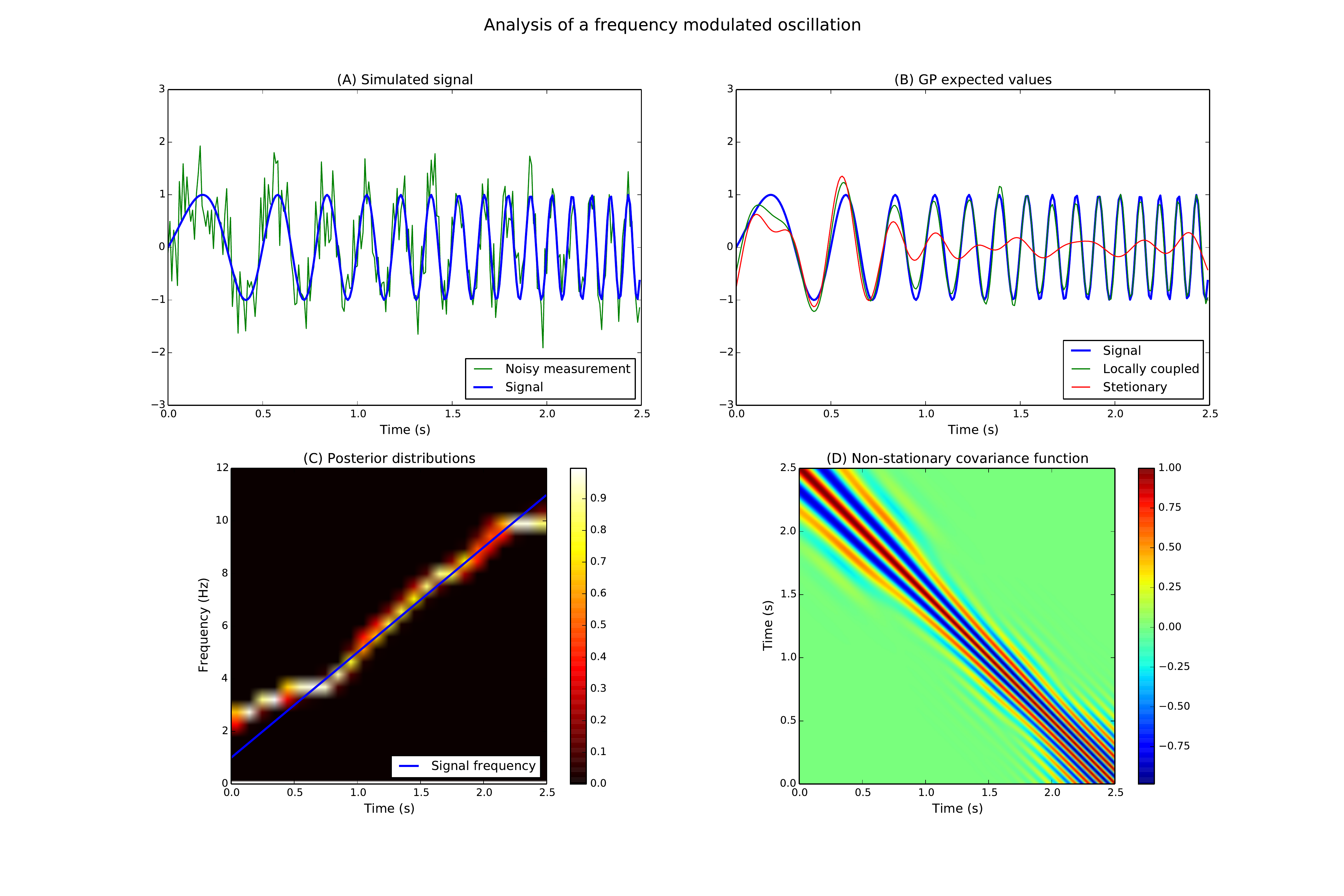}% picture filename
	\caption{Analysis of a frequency modulated oscillation. A) Simulated signal (blue) and noise corrupted measurement (green). B) Comparison of locally coupled GP expected value (green) and stationary GP expected value (red). C) Posterior distribution of the instantaneous frequency. D) Estimated nonstationary covariance function.}
	\label{figure 1}
\end{figure}

As an example of a nonstationary oscillation, we analyzed the following deterministic signal:
\begin{equation}
g(t) = \sin(2 \pi t + 4 \pi t^2)~.
\label{oscillatory test signal, experiments}
\end{equation}
The signal was corrupted with additive Gaussian white noise with standard deviation equal to $0.5$ (SNR = 2). Figure \ref{figure 1}A shows the simulated signal (blue) and the noise corrupted simulated measurement (green). 

We compared the performance of the locally coupled GP with a stationary GP regression equipped with an oscillatory covariance function (see Eq.~\ref{oscillatory, methods}). In this stationary analysis, the width parameter $d$ was kept fixed at $0.4$ s while the peak frequency was estimated from the data by maximizing the model marginal likelihood. Figure \ref{figure 1}B shows the original signal (blue) together with the posterior expected value of the stationary GP regression (red) and the locally coupled GP (green). The locally coupled analysis is able to properly denoise the whole time series while the stationary analysis can only capture a limited segment. Figure \ref{figure 1}C shows the posterior expectations of the hidden variables of the locally coupled GP analysis compared with the real instantaneous frequency (blue line). Finally, figure \ref{figure 1}D shows the resulting global covariance function of the process.

\subsection{Analysis of a Time Series with Two Dynamical States}
Nonlinear physical, biological and chemical systems can switch between several states characterized by different dynamical properties \citep{feudel2008complex}. For example, a neuron can act as either an integrator or an oscillator and can have transitions between these two dynamical states \citep{izhikevich2007dynamical}. As another example, in the human EEG signal, alpha oscillations (7-13 Hz) exhibit transitions between a high amplitude and a low amplitude state \citep{roberts2015heavy}. 

Using the locally coupled GP regression, we can model this kind of phenomena with a set of covariance functions that have different dynamical proprieties. In this case the hidden variable is discrete and specifies which covariance function is active in a local signal segment. Here we will limit our attention to a model with two states, one oscillatory and another broadband. In this case we can use local covariance functions of the following form:
\begin{equation}
k_i(t,t') = \alpha_i k_{SE}(t,t') + (1 - \alpha_i) k_{O}(t,t')~,
\label{multistate covariance, experiments}
\end{equation}
where $\alpha_i$ is a binary variable that can be either zero or one. The distribution over $\alpha_i$ is specified by a $2 \times 2$ transition matrix. 

\begin{figure}
	\centering
    	\includegraphics[width=0.9\textwidth] {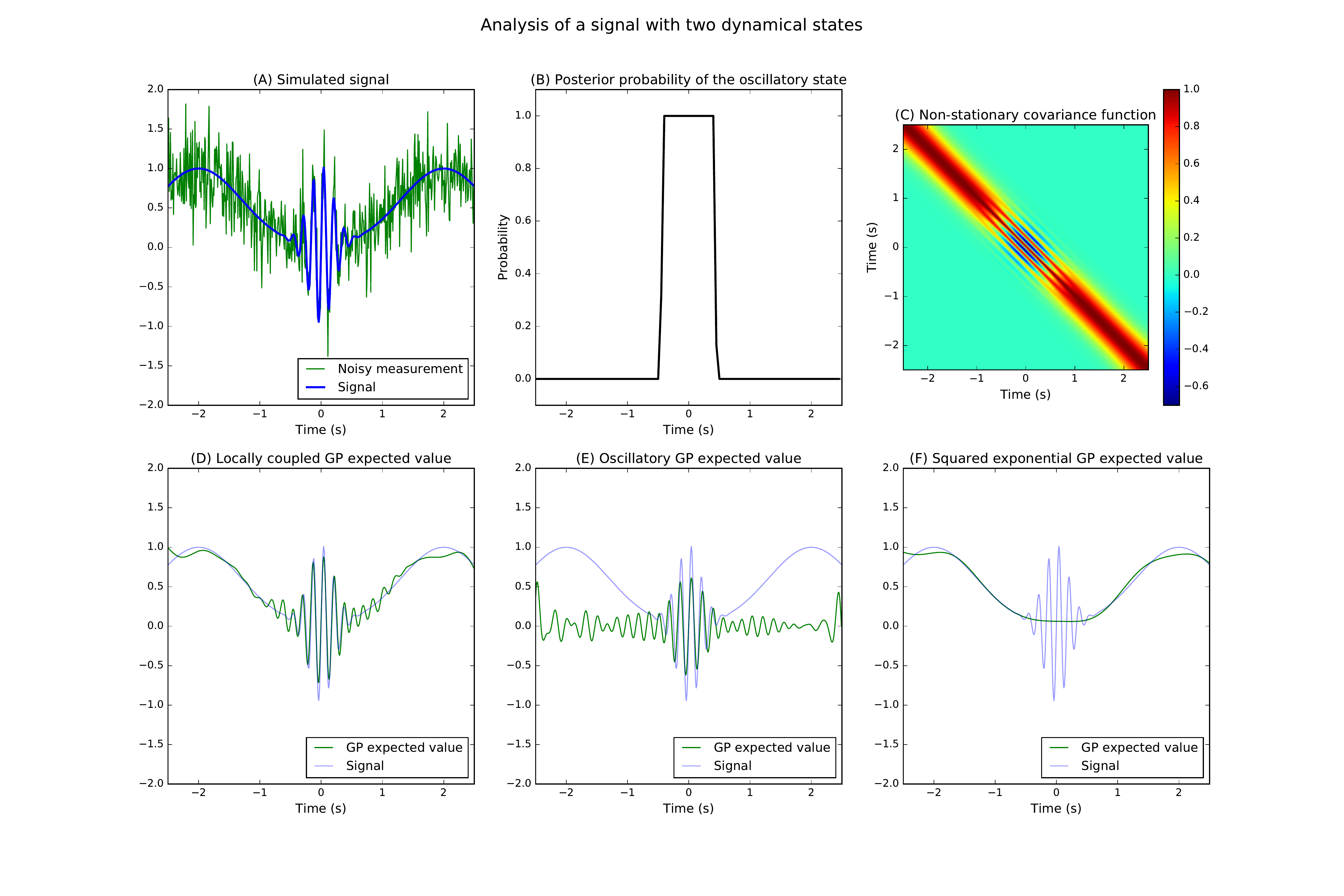}% picture filename
	\caption{Analysis of a signal with two dynamical states. A) Simulated signal (blue) and noise corrupted measurement (green). B) Posterior probability of the oscillatory state as a function of time. C) Estimated nonstationary covariance function. D,E,F) Expected values of locally coupled GP, stationary oscillatory GP and stationary squared exponential GP analysis respectively.}
	\label{figure 2}
\end{figure}

In order to validate this model, we generated the following deterministic signal:
\begin{equation}
f(t) = e^{-\frac{(t + 2)^2}{2 \cdot 0.7^2}} + e^{-\frac{t^2}{2 \cdot 0.3^2}} \cos{2 \pi 5 t} + e^{-\frac{-(t - 2)^2}{2 \cdot 0.7^2}}~,
\label{multistate test signal, experiments}
\end{equation}
with $t$ ranging from $-2.50$ to $2.5$ in steps of $0.01$. We corrupted the signal with additive Gaussian white noise with standard deviation equal to $0.3$ (SNR = 9). Figure \ref{figure 2}A shows the simulated signal (blue) and the noise corrupted simulated measurement (green). 

The simulated measurement was analyzed using the locally coupled GP regression specified by the covariance function \ref{multistate covariance, experiments}. We used Gaussian window functions with a width (standard deviation) of $0.4$ seconds whose centers were spaced by $0.05$ seconds. In the analysis, both the squared exponential and the oscillatory covariance function had a width of $0.04$ seconds, while the frequency of the oscillatory covariance function was equal to $5$ Hz. Figure \ref{figure 2}B shows the posterior probability of $\alpha_i = 1$ for each window center $t_i$, i.e. the probability of the signal being oscillatory as a function of time. Figure \ref{figure 2}C shows the resulting nonstationary covariance function obtained by using the mode of this distribution as point estimator. 

We compared the expected value of the locally coupled GP regression with those of two stationary GP regressions, one with the sole square exponential and the other with the sole oscillatory covariance function. Figures \ref{figure 2}D,E\&F show the posterior expected values of these three methods. The locally coupled GP regression captures both the oscillatory and the broadband part of the signal. Conversely, the stationary oscillatory GP regression fails to capture the slow component and interprets some of the noise as oscillatory activity. Finally, the stationary squared exponential GP regression is able to properly reconstruct the slow component but it completely ignores the oscillatory part. The correlation between the posterior expectation of these three methods and the real signal is respectively 0.97, 0.33 and 0.88.

\subsection{Analysis of Brain Oscillations}
We applied the two-state locally coupled GP analysis on MEG recordings in order to identify bursts of alpha brain oscillations. We collected resting state brain activity from an experimental participant that was instructed to fixate on a cross at the center of a black screen. Brain activity was recorded using a 275 axial gradiometer MEG setup (VSM/CTF Systems, Port Coquitlam, British Columbia, Canada). The present analysis was carried out on six seconds of signal from an occipital sensor. This signal was high pass filtered ($2$ Hz) and subsequently down-sampled at $300$ Hz.

We used a two-state covariance function of the following form:
\begin{equation}
k_i(t,t') = \alpha_i k_{E}(t,t') + (1 - \alpha_i) k_{O}(t,t')~,
\label{multistate covariance II, experiments}
\end{equation}
where the exponential covariance function $k_{E}(t,t')$ determines a continuous but not differentiable stochastic process. This is a suitable phenomenological model for the non-oscillatory brain activity \citep{ambrogioni2016dynamic}. The covariance functions are defined as follows:
\begin{equation}
k_{E}(t,t') = e^{-|t' - t|/k}~.
\label{exponential covariance, experiments}
\end{equation}
The parameter $k$ determines the time scale of the process and was set to $0.3$. The oscillatory covariance function had a width of $0.03$ seconds and its peek frequency was equal to $10$ Hz. Finally, we assumed an SNR of $2$.

\begin{figure}
	\centering
    	\includegraphics[width=0.9\textwidth] {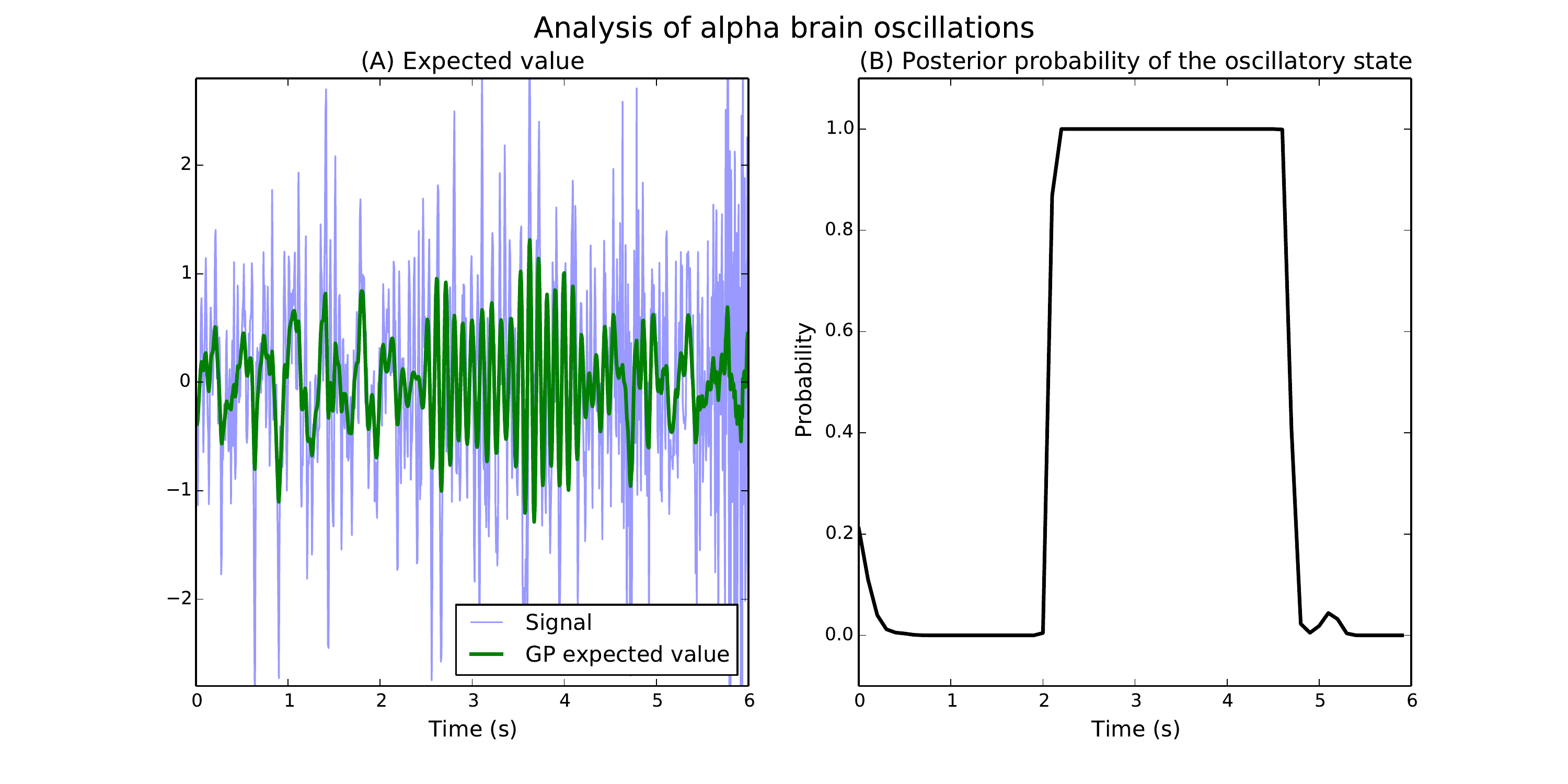}% picture filename
	\caption{Analysis of alpha brain oscillations. A) Measured brain signal (blue) and expected value of the two-state locally coupled GP regression. B) Posterior probability of the oscillatory state as a function of time. }
	\label{figure 3}
\end{figure}

Figure \ref{figure 3} shows the results of this analysis.  Panel A shows the expected value of the coupled GP analysis (green) together with the raw signal (blue). A burst of alpha oscillation is visible from $2$ s to $4.5$ s while the rest of the signal seems to be characterized by more irregular activity. This behavior is captured by the posterior distribution of the hidden variables $\alpha_i$, as shown in panel B.

\section{Conclusions}
In this paper we proposed a new framework for nonstationary GP regression. This framework allows to easily design nonstationary covariance functions whose time-dependent parameters are coupled through a Markov chain. When the range of the parameters is finite, the posterior distribution can be efficiently obtained in closed form using the forward-backward algorithm. 

As example applications, we analyzed two types of nonstationary oscillations. In the first example, the frequency of the oscillation changed as a function of time. We showed that the posterior distribution of the model frequency reliably tracks the frequency shift of the signal. The resulting nonstationary covariance function greatly outperforms the optimized stationary covariance function in denoising the simulated signal. In the second example, we simulated a signal with two distinct dynamical states. On a simulated signal, we showed that the two-state locally coupled GP regression can identify oscillatory bursts amid periods of broadband activity. We further showed that this technique can be applied to identify bursts of alpha oscillations in the human MEG signal.

\bibliography{locally_coupled_GPbiblio}

\end{document}